\documentclass[runningheads]{llncs}

\usepackage{eccv}

\usepackage{colortbl}

\usepackage{eccvabbrv}

\usepackage{graphicx}
\usepackage{booktabs}

\usepackage[accsupp]{axessibility}  

\definecolor{cvprblue}{rgb}{0.21,0.49,0.74}
\usepackage{booktabs}
\usepackage{array}

\usepackage{makecell}
\usepackage{bbding}
\usepackage{utfsym}
\usepackage{pifont}
\usepackage{diagbox}
\usepackage[ruled]{algorithm2e}
\usepackage{wrapfig}

\usepackage{multirow}
\usepackage{amssymb} 
\definecolor{lightblue}{rgb}{0.68, 0.85, 0.90}

\usepackage{hyperref}

\usepackage{orcidlink}

\begin{document}

\title{Adaptive Multi-modal Fusion of Spatially Variant Kernel Refinement with Diffusion Model for Blind Image Super-Resolution} 

\titlerunning{SSR}



\author{Junxiong Lin\inst{1} \and
Yan Wang\inst{1*} \and 
Zeng Tao\inst{1} \and
Boyang Wang\inst{1} \and
Qing Zhao\inst{1} \and \\
Haorang Wang\inst{1} \and 
Xuan Tong\inst{1} \and
Xinji Mai\inst{1} \and
Yuxuan Lin\inst{2} \and \\
Wei Song\inst{3} \and
Jiawen Yu\inst{1} \and
Shaoqi Yan\inst{1} \and 
Wenqiang Zhang\inst{4,5}\thanks{Corresponding author}}
\authorrunning{J.~Lin et al.}

\institute{Shanghai Engineering Research Center of AI \& Robotics, Academy for Engineering \& Technology, Fudan University, Shanghai, China. \and 
East China University of Science and Technology, Shanghai, China. \and
College of Information Technology, Shanghai Ocean University, Shanghai, China. \and
Engineering Research Center of AI \& Robotics, Ministry of Education, Academy for Engineering \& Technology, Fudan University, Shanghai, China. \and
Shanghai Key Lab of Intelligent Information Processing, School of Computer Science, Fudan University, Shanghai, China \\
\email{\{linjx23\}@m.fudan.edu.cn} \space
\email{\{yanwang19,wqzhang\}@fudan.edu.cn}
}
\maketitle
\begin{abstract}
  Pre-trained diffusion models utilized for image generation encapsulate a substantial reservoir of a priori knowledge pertaining to intricate textures. Harnessing the potential of leveraging this a priori knowledge in the context of image super-resolution presents a compelling avenue. Nonetheless, prevailing diffusion-based methodologies presently overlook the constraints imposed by degradation information on the diffusion process. Furthermore, these methods fail to consider the spatial variability inherent in the estimated blur kernel, stemming from factors such as motion jitter and out-of-focus elements in open-environment scenarios. This oversight results in a notable deviation of the image super-resolution effect from fundamental realities. To address these concerns, we introduce a framework known as Adaptive Multi-modal Fusion of \textbf{S}patially Variant Kernel Refinement with Diffusion Model for Blind Image \textbf{S}uper-\textbf{R}esolution (SSR). Within the SSR framework, we propose a Spatially Variant Kernel Refinement (SVKR) module. SVKR estimates a Depth-Informed Kernel, which takes the depth information into account and is spatially variant. Additionally, SVKR enhance the accuracy of depth information acquired from LR images, allowing for mutual enhancement between the depth map and blur kernel estimates. Finally, we introduce the Adaptive Multi-Modal Fusion (AMF) module to align the information from three modalities: low-resolution images, depth maps, and blur kernels. This alignment can constrain the diffusion model to generate more authentic SR results. 
\end{abstract}

\begin{figure}[th]
  \centering
    \includegraphics[width=\linewidth]{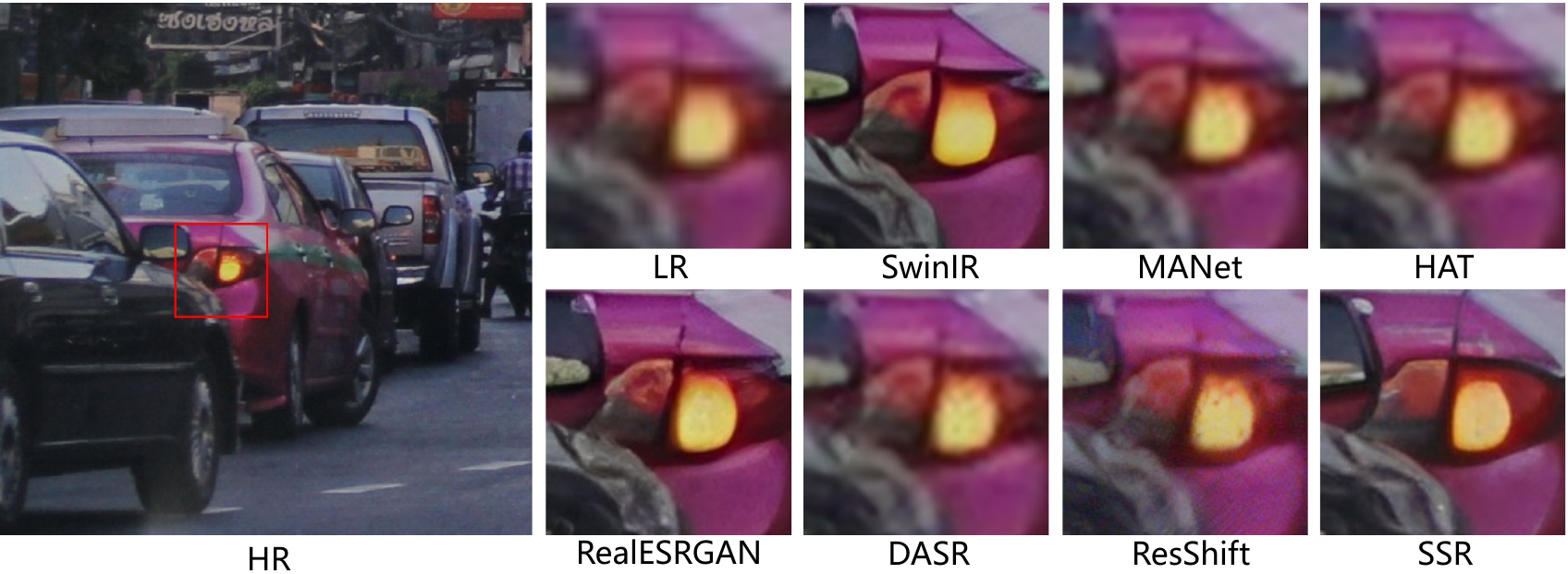}

   \caption{Visual comparison (×4) on DRealSR. SwinIR, MANet, HAT, RealESRGAN, DASR and ResShift suffer from noise and blurring artifacts, while SSR can generate high-fidelity images.}
   \label{fig:fig1}
\end{figure}

\section{Introduction}
\label{sec:intro}

Image super-resolution is a foundational pursuit within the domain of computer vision, focusing on the reconstruction of high-resolution images from their low-resolution counterparts to significantly augment pixel density \cite{chen_liang_zhang_liu_zeng_zhang_2023, chao_zhou_gao_gong_yang_zeng_dehbi_0}. This intricate task involves the inference and restoration of high-frequency details that are often obscured within the constraints of limited pixel resolutions. Mathematically, the regression process from a high-resolution image (HR) to a low-resolution image (LR) is modeled as follows:\begin{equation}
I^{LR}=(I^{HR}\otimes k)\downarrow_s+n \label{eq:srdown}
\end{equation}where $k$ symbolizes the blur kernel, $\otimes$ denotes the convolution operation, $\downarrow_s$ signifies a downsampling by a factor of $s$, $n$ represents noise. Over the past decade, methods grounded in CNN or transformers have made remarkable strides \cite{chen_zhang_gu_kong_yang_yu_2023, choi_lee_yang_2022, lutio_becker_d’aronco_russo_wegner_schindler_0,zhang2012two}. Traditional super-resolution methods assume this degradation is characterized by a specific blur kernel, such as a bicubic kernel \cite{magid_lin_wei_zhang_gu_pfister_university_college_zürich_0}. However, when a well-trained SR model receives LR images away from its training distribution, it often suffers significant performance degradation \cite{li_fan_xiang_demandolx_ranjan_timofte_gool_2023}. To tackle this problem, blind super-resolution tasks aim at image super-resolution with unknown degradation processes \cite{liu_cheng_tan_0}.

\begin{figure}[t]
  \centering
    \includegraphics[width=\linewidth]{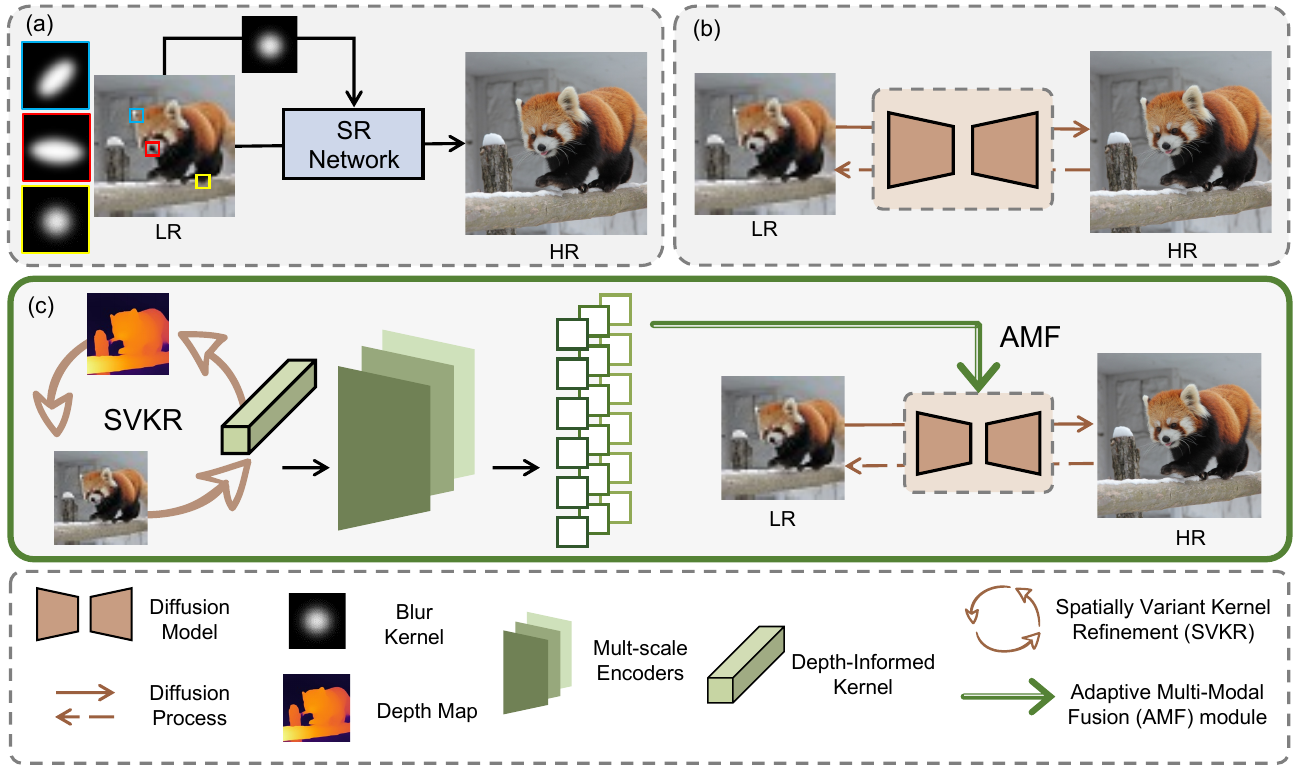}

   \caption{The illustration of various blind super-resolution methods. (a) The majority of super-resolution methods assume that the image degradation process is spatially invariant, estimating only a single blur kernel for an individual image. (b) Diffusion based super-resolution methods using texture prior information. (c) Our SSR approach, which imposes constraints on the diffusion solution space through spatially variant blur kernels and depth information.}
   \label{fig:intro}
\end{figure}

As depicted in Figure \ref{fig:intro}, spatially invariant estimated blur kernels, incorporating a priori degradation information, are employed in the super-resolution process to enhance performance and achieve a finite solution space for the LR-HR image pair \cite{cao_wang_xian_li_ni_pi_zhang_zhang_timofte_gool_2022,chen_xu_hong_tsai_kuo_lee_2023}. In real-world imaging, factors limiting image resolution extend beyond specific blur kernels to encompass motion blur, depth variation in imaging objects, and optical characteristics of imaging devices \cite{conf/cvpr/XuWCLMLL22,yan2015software,jiang2023efficient,zhang2023query}. The use of a spatially invariant blur kernel becomes inadequate in accurately describing this multifaceted process, yielding unsatisfactory results post super-resolution \cite{yue_zhao_xie_zhang_meng_wong_0}. Some researchers address this by leveraging a priori knowledge from a pre-trained diffusion model \cite{zheng_zhou_li_qi_shan_li_0}, compensating for lost texture information in LR images. However, the lack of information constraints coupled with the inherent randomness of the diffusion model contributes to the deviation from fundamental principles.

To address these challenges, we introduce the SSR framework, leveraging depth information to guide blur kernel estimation and subsequently inform the diffusion process \cite{ wei_zhang_2023,yuan2024lcseg}. This approach facilitates HR image reconstruction from LR images without compromising the diffusion prior inherent in the generative model. For spatially variant blur kernel estimation, we proposed the Spatially Variant Kernel Refinement (SVKR) module. Within SVKR, Depth-Informed Kernel Estimate Network (DKENet) is devised to utilize the depth information across image locations, and estimate spatially variant blur kernels based on depth information. Introducing depth information enhances the variability of depth information for different objects, aiding in characterizing target object contours. Additionally, SVKR further refines the accuracy of estimated kernel and depth information from blurred, low-resolution images. Moreover, to fully exploit the value of depth information, we have developed the Adaptive Multi-Modal Fusion (AMF) module to merge information from multiple modalities, thereby constraining the diffusion model with spatially variant degradation.


In summary, our main contributions are four-fold: 1) We propose a blind image super-resolution framework known as SSR. SSR employs multiple scales of encoders for information from different modalities using the AMF module, thus generates images with higher fidelity without compromising the diffusion priors. 2) We proposed a Spatially Variant Kernel Refinement (SVKR) module to estimate blur kernels. SVKR iteratively enhance the LR images, thereby refining the precision of blur kernel and depth information estimation. 3) We developed Adaptive Multi-Modal Fusion (AMF) module to merge information from multiple modalities, thereby constraining the diffusion model to generate more authentic SR results. 4) Extensive quantitative and qualitative experiments on representative datasets have verified the superior performance and effectiveness of our method.

\section{Related Work}

\subsection{Non-Blind SR}
To enhance non-blind super-resolution, various methods have been developed \cite{conf/cvpr/LuoHYLFL22, kong2021deep,Zhou_2023_ICCV, lin2024suppressing}. SRMD \cite{zhang2018learning} introduced intricate blur kernel assumptions and incorporated them into the network. SFTMD \cite{gu2019blind} fused blur kernels using SFT layers, while DPSR \cite{zhang2019deep} offered a plug-and-play framework for uniform blur kernels. UDVD \cite{Sheth_2021_ICCV} used dynamic convolution to address varying degradations, and DPIR \cite{zhang2021plug} employed a HQS to iteratively address data subproblems and prior subproblems.
Additionally, zero-shot super-resolution methods have emerged. ZSSR \cite{ZSSR} introduced unsupervised CNN super-resolution, and DGDML-SR \cite{cheng2020zero} used depth information for network training. However, these methods face challenges related to foundational assumptions. For example, it is inaccurate for DGML-SR to determine the resolution as high or low solely based on distance. Although these non-blind super-resolution methods have made commendable progress, they often exhibit disappointing performance when applied to complex real-world scenarios.

\subsection{Kernel Estimate}
In most blind super-resolution approaches, degradation characteristics from LR images are integrated with non-blind super-resolution methods. IKC \cite{gu2019blind} improves accuracy through iterative approaches and corrective function design. DCLS \cite{luo2022deep} redefines degradation for low-resolution space deblurring, and DASR \cite{Wang2021Unsupervised} uses contrastive learning for degradation characterization. However, these methods typically estimate a single blur kernel or degradation feature for the entire image, causing disparities in spatially variant degradation domains and performance degradation in practical applications.

Efforts to address domain disparities include estimating spatially variant blur kernels. KOALAnet \cite{kim_sim_kim_2021} adapts kernels locally for collaborative learning. MANet \cite{liang2021mutual} maintains degradation locality with a moderately sized receptive field but limits global information. CMOS \cite{chen_zhang_xu_wang_wang_liu_2023} simultaneously estimates semantic and blur maps, integrating cross-modal features for defocused image restoration but faces limitations in addressing various degradation processes.

\subsection{Diffusion Model Based SR}
Denoising Diffusion Probability Models (DDPM) \cite{Ho_Jain_Abbeel_Berkeley} have introduced a new generation paradigm, leveraging Markov chains to create stable inverse operations for complex generative processes. Researchers have used DDPM for image SR \cite{gao_liu_zeng_xu_li_luo_liu_zhen_zhang_2023, li_zuo_loy_2023}. SR3 \cite{Saharia_Ho_Chan_Salimans_Fleet_Norouzi_2022} adopts a supervised diffusion model-based approach, coupling degraded images with generated images to facilitate conditional image generation for SR. In contrast, ILVR \cite{Choi_Kim_Jeong_Gwon_Yoon_2021} replaces the low-frequency components in the denoising output with those from the reference image, utilizing an unconditional diffusion model for no-training conditional generation in image SR. CDPMSR \cite{niu2023cdpmsr} improves low-quality images by leveraging pre-existing SR models, providing better conditions for the diffusion model.

Given the intrinsic demand for high fidelity in SR, generated images must maintain a strong consistency with real images. However, these methods employ LR images or enhanced LR images as the diffusion conditions, resulting in a disparity between the images generated in the sampling space of the diffusion model and real images \cite{li2023diffusion}. Our approach introduces depth maps and blur kernels as additional conditioning information, thereby constraining the sampling space, and rendering the generated images more closely aligned with real images.
\begin{figure*}[t]
  \centering
    \includegraphics[width=\linewidth]{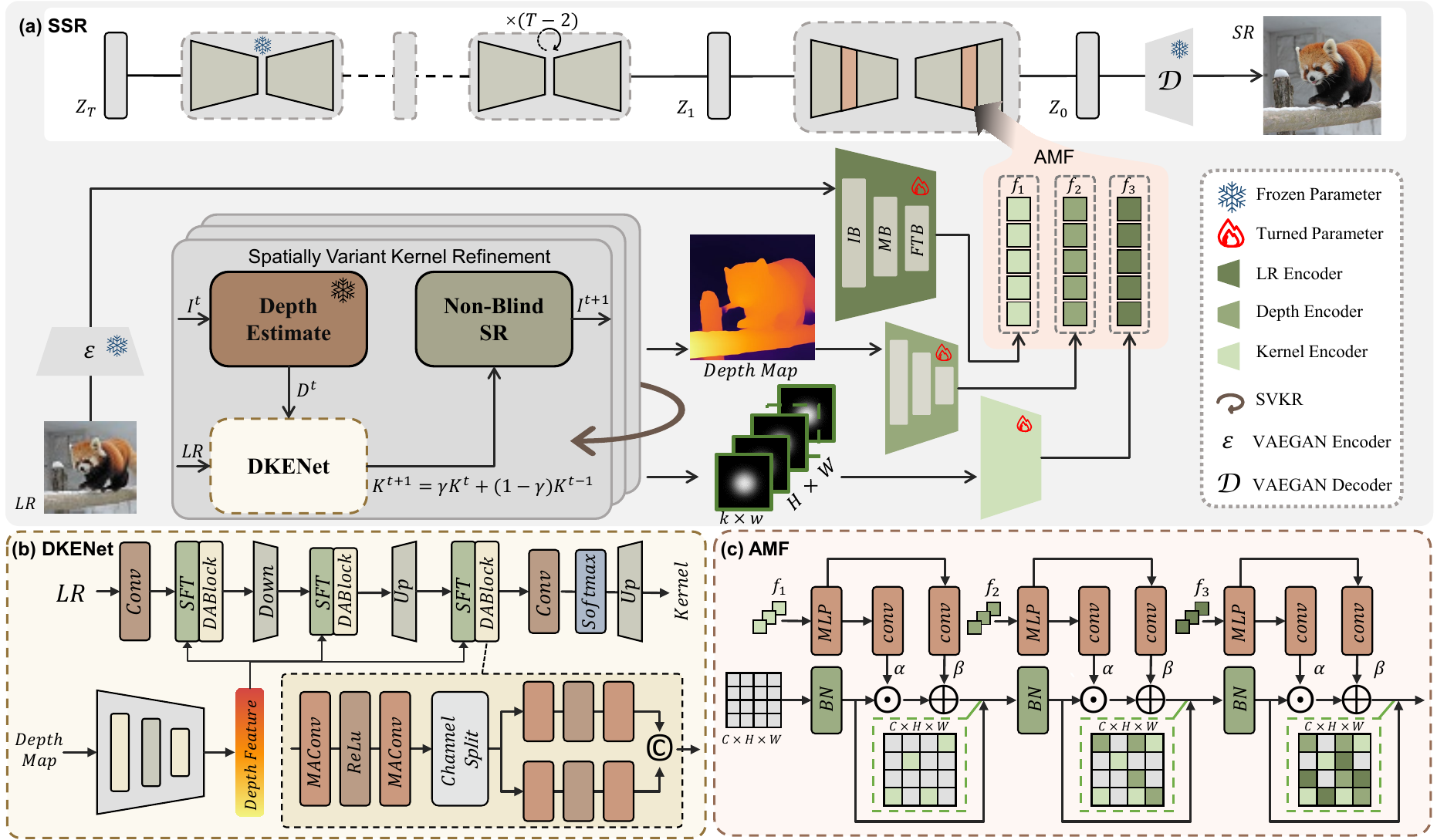}

   \caption{The framework of the proposed Adaptive Multi-modal Fusion of Spatially Variant Kernel Refinement with Diffusion Model for Blind Image Super-Resolution (SSR). (a) Illustration of the main process of SSR. (b) Depiction of the Depth-Informed Kernel Estimate Network (DKENet) for spatially variant kernel estimation. (c) Depiction of the Adaptive Multi-Modal Fusion (AMF) module for information fusion.}
   \label{fig:frame}
\end{figure*}
\section{Methodology}

\subsection{Overview}
As previously stated, the application of a spatially variant kernel and depth information can significantly enhance the accuracy of image SR tasks. With the development of deep learning, many network architectures have emerged, including CNN \cite{taoMM, wang2024seeking}, Transformer \cite{tao20243, mai2024ous}, and diffusion model \cite{Ho_Jain_Abbeel_Berkeley,mai2024efficientmultimodalmodelsworld,gao_liu_zeng_xu_li_luo_liu_zhen_zhang_2023}. Pre-trained diffusion models encapsulate an extensive repository of a priori knowledge concerning intricate textures. Drawing inspiration from this premise, we introduce a framework named Adaptive Multi-modal Fusion of Spatially Variant Kernel Refinement with Diffusion Model for Blind Image Super-Resolution (SSR). Through the integration of multi-modal data, we impose constraints on the diffusion model's sampling space, thereby guiding the diffusion process to produce images that closely resemble reality.

Our initial step involves the iterative estimation of depth information and spatially variant blur kernels from low-resolution images (LR) with DKENet. Specifically, within the iterative estimation process, we first infer the depth map and estimate the spatially variant blur kernels. Subsequently, we combine the blur kernel with a non-blind SR method based on RRDB \cite{wang2018esrgan} to estimate a high-resolution image. From this high-resolution image, we once again estimate the depth information. Iterating this process ensures that the estimation of depth information and the blurred kernel mutually reinforce each other, thereby enhancing the precision of the guidance information. Meanwhile, inspired by StableDiffusion \cite{rombach2021highresolution}, we encode LR images into a latent space using a VAEGAN encoder \cite{larsen2016autoencoding}, and this encoded information guides the diffusion model. These process produce three modalities of LR image, Depth map and blur kernel. Subsequently, information from three different modalities is encoded using encoders of varying scales and is aligned through an Adaptive Multi-Modal Fusion (AMF) module. This aligned information is employed to guide the diffusion process. Through AMF, the aligned features can effectively fine-tune the diffusion model, reducing the number of parameters and the complexity of training. The overall architectural diagram is depicted in Figure \ref{fig:frame}.
\subsection{Spatially Variant Kernel Refinement} 

\noindent \textbf{Depth-Informed Kernel Estimate Network.} In the realm of blind SR, the high-resolution image $x$ and the kernel $k$ in Equation \eqref{eq:srdown} are unknown. We need to estimate $k$ and then use it as prior knowledge to reconstruct $x$. For an image $x \in \mathcal{R}^{H \times W}$, we undertake the estimation of a kernel $k \in \mathcal{R}^{hw \times H \times W}$, where $h$ and $w$ respectively denote the height and width of the kernel. Each component of $k$ corresponds to an individual pixel within $x$.

We utilize Depth-Informed Kernel Estimate Network to estimate the Depth-Informed Kernel. The Depth-Informed Kernel is a spatially variant blur kernel that incorporates depth information of the image during its estimation process. DKENet is structured into two distinct branches, each addressing a separate input modality: one branch processes LR images, while the other extracts features from depth maps. In the LR image processing branch, the network's operation commences with a standard $3 \times 3$ convolutional layer, purposed for preliminary feature extraction. Subsequently, the input features are channeled into a Depth-Aware Block, enhancing the network's responsiveness to pivotal features. 

Specifically, within the Depth-Aware Block, features are initially merged with depth map information through a simple yet widely applied SFT layer \cite{gu2019blind}, followed by two MAconv layers \cite{liang2021mutual} interspersed with ReLU activations. To enable the network to distinctively explore useful features and acquire more nonlinear representational capabilities, we integrated a Channel Split layer, bifurcating the channel count and separately extracting features before their recombination. 

Following this, the feature map undergoes a downsampling operation to reduce its spatial resolution, and then passes through another DABlock for further feature refinement. Subsequent upsampling increases the resolution of the feature map, ensuring the density of information. After another pass through a DABlock, the features proceed through a convolutional layer and culminate in a Softmax layer and a upscale-layer. Regarding the depth map branch, it undergoes a feature extraction process similar to that of the LR image branch, ultimately converging with the LR image branch through an SFT layer.

\noindent \textbf{Spatially Variant Kernel Refinement.} The acquisition of depth information often requires specialized equipment like Time-of-Flight cameras and laser rangefinders \cite{metzger_daudt_schindler_2022, Ranftl_Lasinger_Hafner_Schindler_Koltun_2022}, which are not always practical for image SR applications \cite{song2020enhancement, wang2019experimental}. Thus, we use a monocular depth estimation algorithm to derive depth information from low-resolution images. It's crucial to note that image resolution significantly influences the accuracy of depth estimation; acquiring depth from a blurry, low-resolution image is inherently imprecise \cite{park_moon_cho_2022, tian_lu_liu_guo_chen_zhang_0}. To overcome this, we have devised Spatially Variant Kernel Refinement (SVKR) to augment low-resolution images, subsequently deriving depth information from the resulting images. This intricate process synergizes depth information to facilitate the estimation of the blur kernel.



\begin{algorithm}
\SetAlgoNlRelativeSize{0}

\KwIn{Low-resolution image $LR$, Monocular depth estimation algorithm $DE$, DKENet $KE$, Number of iterations $T$, $\gamma=0.9$}
\KwOut{Depth Map $D$, Kernel $K$}

Initialization: \space
    $t=0$, $I_0 = LR$; \space
    $D^0 = DE(I^0)$; \space
    $K^0 = KE(LR, D^0)$;

\While{$t < T$}{
  Estimate depth map: $D^{t+1} \leftarrow DE(I^t)$;

  Estimate kernel: $K^{t+1} \leftarrow KE(LR, D^{t+1})$; 
  \space $K^{t+1}\leftarrow\gamma K^{t+1}+(1-\gamma)K^{t}$\; \space
    
  Non-blind SR: $I^{t+1} \leftarrow NBSR(LR, K^{t+1})$; \space
  $t \leftarrow t + 1$\;
}

\textbf{Return} Final depth map: $D^T$, Final kernel: $K^T$\;
\caption{Spatially Variant Kernel Refinement}
\end{algorithm}

Algorithm 1 shows the process of SVKR. $DE$ is the current state-of-the-art monocular estimation algorithm, while $KE$ assumes the role of a spatially varying kernel estimation algorithm, and $NBSR$ stands as a lightweight, RRDB-based non-blind SR method. In the course of these iterations, $DE$ estimates the depth map from the most recent SR image $I^{t}$, $KE$ harnesses the most recent depth map $D^{t}$ to estimate the blur kernel from the LR image, and $NBSR$ amalgamates the latest blurred kernel $K^{t}$ to enhance the super-resolved image. This intricate coordination effectively combines the estimations of depth maps and blur kernels, mutually enhancing each other's precision.

\begin{figure*}[t]
  \centering
    \includegraphics[width=\linewidth]{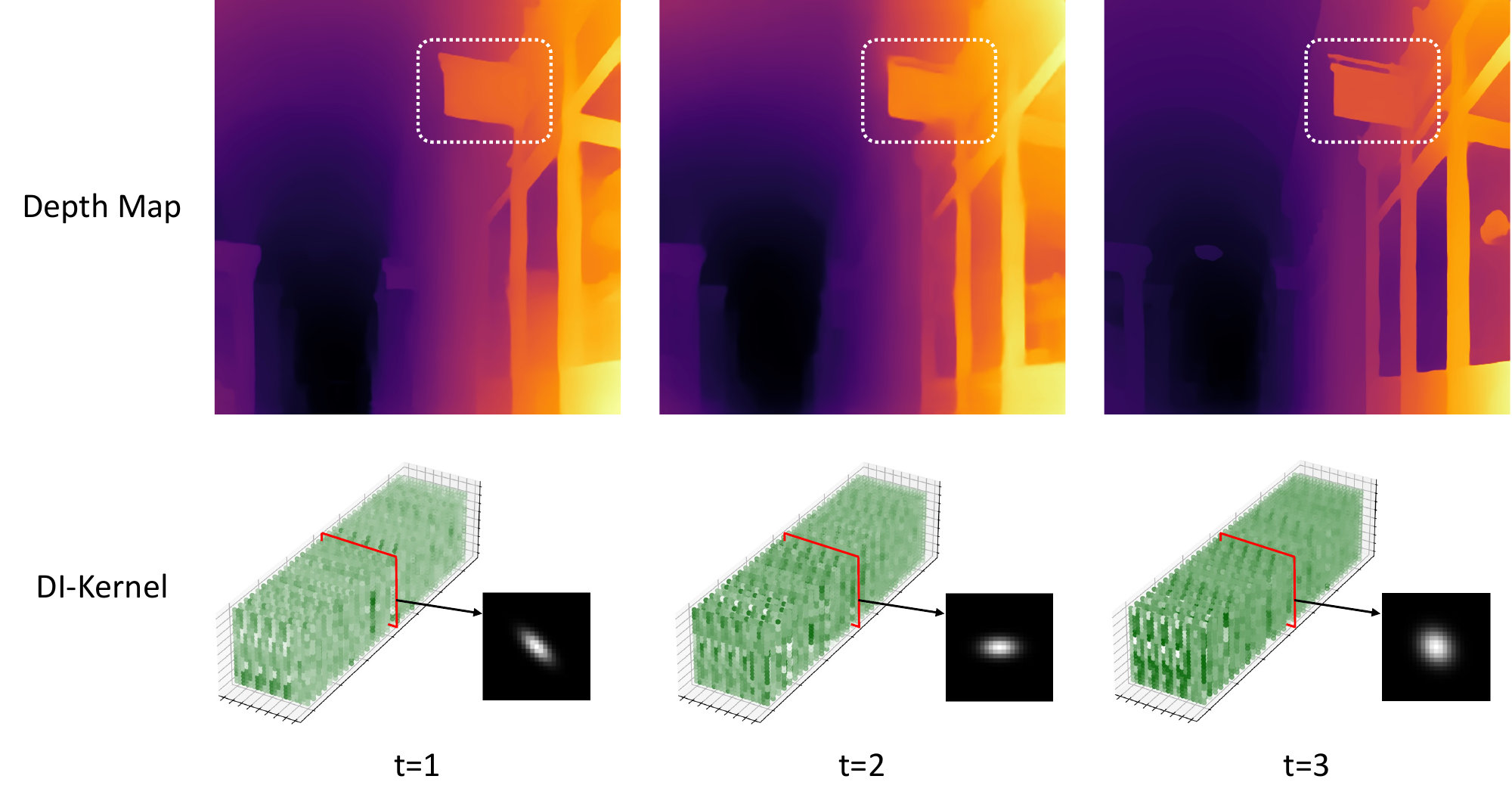}

   \caption{The visualization of depth maps and DI-Kernels during the SVKR iteration process, where depth information and degradation information are mutually enhanced during the iteration process.}
   \label{fig:visual}
\end{figure*}

When $t \geq 1$, $K$ is updated using Equation \eqref{eq:IE}, where $\gamma$ is the hyperparameter that governs the update rate. This approach allows for the retention of the influence of historical information while updating $K$, rendering it persistent in the direction of updates and mitigating the instability of results caused by oscillations during the updating process. Figure \ref{fig:visual} demonstrates the effect of iterative enhancement of depth information and degradation information by SVKR.
\begin{equation}
K^{t+1}=\gamma K^{t}+(1-\gamma)K^{t-1} \label{eq:IE}
\end{equation}

\subsection{Adaptive Multi-Modal Fusion.} Due to the distinct structural and characteristic attributes exhibited by data from various modalities, we have carefully devised encoders of diverse scales and architectures. We use a half UNet as the encoder to encode LR images. This encoder comprises the Input Block (IB), the Middle Block (MB), and the Feature Translate Block (FTB). Among the three modalities of information, namely LR images, depth maps, and blur kernels, LR images serve as the primary modality, containing the most abundant repository of information. Hence, we have tailored the largest-scale encoder to accommodate this primary modality. Conversely, depth maps and blur kernels carry a relatively limited amount of informational content, thus warranting the implementation of smaller-scale encoders. In order to amalgamate the information from these three modalities and employ it for guiding the diffusion process, we have designed the adaptive multi-modal Fusion (AMF) module, as depicted in Equation \eqref{eq:CCFM}, where $i \in \{1,2,3\}$.
\begin{equation}
F^{i}_{out}=(1+M_{\alpha i}(f_i))\odot BN(F^{i}_{in})\oplus M_{\beta i}(f_i) \label{eq:CCFM}
\end{equation}
where $f_1$, $f_2$, and $f_3$ respectively denote the features obtained through the encoding of the blurred kernel, depth map, and LR image. The symbol $BN(\cdot)$ signifies batch normalization, while $M_{\alpha}$ and $M_{\beta}$ denote MLPs. Specifically, $M_{\alpha}$ serves to rescale features along the channels, while $M_{\beta}$ applies an offset to features, ensuring the dynamic adjustment of semantic information for achieving alignment across three modalities in the semantic space. For the LR image denoted as $x$, the objective of the SR diffusion model is to acquire the posterior distribution $p_\theta(z_t-1|x, z_t)$ at a given time step $t$, while satisfying the initial condition $z_0 \sim q(z|x)$, where $z$ represents the corresponding HR image. In accordance with the AMF framework, this objective may be articulated as Equation \eqref{eq:p}.
\begin{equation}
p_\theta(z_{t-1}|(f_1,f_2,f_3),z_t) \label{eq:p}
\end{equation}Thus, the constraints on the diffusion process can be delineated as Equation \eqref{eq:constraints}.
\begin{equation}
L(u)=l(u)+{\lambda}_1C_1(f_1)+{\lambda}_2C_2(f_2)+{\lambda}_3C_3(f_3) \label{eq:constraints}
\end{equation} Where $u$ serves as the output of the diffusion model, $l=\mathbb{E}_{t,z_0,\epsilon}[||\epsilon-\epsilon_\theta(\sqrt{\hat{\alpha}_t}z_0+\epsilon\sqrt{1-\hat{\alpha}_t},t)||_2^2]$ represents the original loss function constraining the diffusion model, $\lambda_i$ corresponds to the scaling factors, and $C_i$ denotes the L1 regularization applied to feature $f_i$. We used L1 regularization because \( f_i \) are features from different modalities, and sparsifying them helps the network learn complementary information.
\section{Experiment}


\begin{table*}[!t]
\centering
\caption{Quantitive results on DIV2K valid, BSDS100, Urban100 and DRealSR datasets for scaling factor $\times 4$. $\uparrow$ indicates that a higher value is better, $\downarrow$ indicates that a lower value is better. \textbf{Bold indicates the best performance.}}  
\vspace{-2mm}
\scalebox{0.78}{
\begin{tabular}{ccccccccc}
\toprule
                           &                          &                            & \multicolumn{6}{c}{Metrics}                                                                                                                                \\
\multirow{-2}{*}{Datasets} & \multirow{-2}{*}{Method} & \multirow{-2}{*}{Backbone} & PSNR $\uparrow$ & SSIM $\uparrow$ & LPIPS $\downarrow$         & MUSIQ $\uparrow$ & CLIP-IQA $\uparrow$ & NIQE $\downarrow$         \\ \midrule
                           & DAN \cite{luo2021endtoend}             & CNN                        & 22.17                       & 0.72                        & 0.42          & 56.98                        & 0.39                            & 6.13          \\
                           & DCLS \cite{luo2022deep}                & CNN                        & 22.41                       & 0.74                        & 0.41          & 54.12                        & 0.39                            & 6.26          \\
                           & DASR \cite{Wang2021Unsupervised}       & CNN                        & 21.45                       & 0.67                        & 0.52          & 54.25                        & 0.45                            & 6.77          \\
                           & MANet \cite{liang2021mutual}           & CNN                        & 18.95                       & 0.60                        & 0.44          & 49.32                        & 0.38                            & 7.30          \\
                           & SwinIR \cite{liang2021swinir}          & Transformer                & 22.08                       & 0.73                        & 0.32          & 68.75                        & 0.57                            & 3.96          \\
                           & HAT \cite{chen2023hat}                 & Transformer                & 22.01                       & 0.72                        & 0.41          & 53.80                        & 0.33                            & 6.31          \\
                           & RealESRGAN \cite{wang2021realesrgan}   & GAN                        & 22.23                       & 0.73                        & 0.32          & 69.41                        & 0.57                            & 4.44          \\
                           & DDNM \cite{wang2022zero}   & Diffusion Model                        & 21.50                       & 0.72                        & 0.37          & 64.22                        & 0.62                            & 3.99          \\
                           & ResShift \cite{yue2023resshift}        & Diffusion Model                       & 22.38                       & 0.72                        & 0.31          & 70.58                        & 0.67                            & 5.16          \\ 
                           \rowcolor{lightblue}
\multirow{-12}{*}{DIV2K \cite{Agustsson_2017_CVPR_Workshops}}    &  SSR(Ours)                                    & Diffusion Model                       & \textbf{22.77}              & \textbf{0.75}               & \textbf{0.30} & \textbf{71.44}               & \textbf{0.69}                   & \textbf{3.82} \\ \midrule \midrule
                           & DAN \cite{luo2021endtoend}             & CNN                        & 27.95                       & 0.87                        & 0.30          & 47.46                        & 0.43                            & 7.03          \\
                           & DCLS \cite{luo2022deep}                & CNN                        & 28.03                       & 0.89                        & 0.27          & 45.07                        & 0.45                            & 7.19          \\
                           & DASR \cite{Wang2021Unsupervised}       & CNN                        & \textbf{28.79}              & 0.89                        & 0.28          & 46.12                        & 0.49                            & 7.43          \\
                           & MANet \cite{liang2021mutual}           & CNN                        & 22.79                       & 0.77                        & 0.31          & 41.10                        & 0.40                            & 8.05          \\
                           & SwinIR \cite{liang2021swinir}          & Transformer                & 26.08                       & 0.84                        & 0.31          & 63.99                        & 0.56                            & 4.54          \\
                           & HAT \cite{chen2023hat}                 & Transformer                & 28.04                       & 0.87                        & 0.29          & 43.82                        & 0.37                            & 7.17          \\
                           & RealESRGAN \cite{wang2021realesrgan}   & GAN                        & 26.62                       & 0.85                        & 0.29          & 62.85                        & 0.52                            & 4.70          \\
                           & DDNM \cite{wang2022zero}   & Diffusion Model                        & 29.48                       & 0.87                        & 0.34          & 27.12                        & 0.32                            & 8.45          \\
                           & ResShift \cite{yue2023resshift}        & Diffusion Model                       & 26.40                       & 0.81                        & 0.38          & 65.35                        & 0.69                            & 5.90          \\ 
                           \rowcolor{lightblue}
\multirow{-12}{*}{BSDS100 \cite{amfm_pami2011}}  & SSR(Ours)                                    & Diffusion Model                       & 27.13                       & \textbf{0.90}               & \textbf{0.25} & \textbf{68.86}               & \textbf{0.72}                   & \textbf{3.78} \\ \midrule \midrule
                           & DAN \cite{luo2021endtoend}             & CNN                        & 20.26                       & 0.69                        & 0.40          & 56.80                        & 0.41                            & 6.15          \\
                           & DCLS \cite{luo2022deep}                & CNN                        & 21.18                       & 0.72                        & 0.37          & 55.07                        & 0.40                            & 6.28          \\
                           & DASR \cite{Wang2021Unsupervised}       & CNN                        & 21.36                       & 0.73                        & 0.38          & 55.17                        & 0.43                            & 6.72          \\
                           & MANet \cite{liang2021mutual}           & CNN                        & 17.05                       & 0.54                        & 0.40          & 52.56                        & 0.38                            & 6.79          \\
                           & SwinIR \cite{liang2021swinir}          & Transformer                & 20.37                       & 0.71                        & 0.29          & 71.76                        & 0.65                            & 4.28          \\
                           & HAT \cite{chen2023hat}                 & Transformer                & 19.56                       & 0.67                        & 0.39          & 56.12                        & 0.38                            & 6.33          \\
                           & RealESRGAN \cite{wang2021realesrgan}   & GAN                        & 20.61                       & 0.72                        & 0.29          & 71.92                        & 0.61                            & \textbf{4.10} \\
                           & DDNM \cite{wang2022zero}   & Diffusion Model                        & 20.01                       & 0.73                        & 0.29          & 51.65                        & 0.24                            & 5.94          \\
                           & ResShift \cite{yue2023resshift}        & Diffusion Model                       & \textbf{21.71}              & \textbf{0.74}               & 0.27          & 71.70                        & 0.68                            & 5.50          \\ 
                           \rowcolor{lightblue}
\multirow{-12}{*}{Urban100 \cite{Huang_CVPR_2015}} & SSR(Ours)                                    & Diffusion Model                       & 21.23                       & 0.71                        & \textbf{0.27} & \textbf{72.74}               & \textbf{0.70}                   & 4.68          \\  \midrule \midrule
                           & DAN \cite{luo2021endtoend}             & CNN                        & \textbf{31.00}                       & 0.92                        & 0.39          & 25.24                        & 0.38                            & 9.06          \\
                           & DCLS \cite{luo2022deep}                & CNN                        & 30.97                       & 0.92                        & 0.41          & 24.71                        & 0.41                            & 10.26        \\
                           & DASR \cite{Wang2021Unsupervised}       & CNN                        & 30.98              & 0.92                        & 0.41          & 27.07                        & 0.40                            & 8.88          \\
                           & MANet \cite{liang2021mutual}           & CNN                        & 27.42                       & 0.90                        & 0.41          & 24.67                        & 0.39                            & 9.58          \\
                           & SwinIR \cite{liang2021swinir}          & Transformer                & 28.45                       & 0.89                        & 0.36          & 51.95                        & 0.40                            & 5.31          \\
                           & HAT \cite{chen2023hat}                 & Transformer                & 30.96                       & \textbf{0.93}                        & 0.41          & 23.66                        & 0.41                            & 10.37          \\
                           & RealESRGAN \cite{wang2021realesrgan}   & GAN                        & 29.94                       & 0.92                        & 0.35          & 47.78                        & 0.38                            & 5.20          \\
                           & DDNM \cite{wang2022zero}   & Diffusion Model                        & 25.01                       & 0.73                        & 0.49          & 41.65                        & 0.44                            & 5.95          \\
                           & ResShift \cite{yue2023resshift}        & Diffusion Model                       & 25.77                       & 0.71                        & 0.57          & 43.23                        & 0.57                            & 9.09          \\ 
                           \rowcolor{lightblue}
\multirow{-12}{*}{DRealSR \cite{wei2020component}}  & SSR(Ours)                                    & Diffusion Model                       & 28.33                       & 0.85               & \textbf{0.30} & \textbf{63.58}               & \textbf{0.64}                   & \textbf{5.01} \\ \bottomrule
\bottomrule
\end{tabular}
} 
\label{tab: general}
\end{table*}

\subsection{Experiment Setup}
\noindent \textbf{Implementation details.}
In accordance with the prevailing blind SR methodology, DKENet employed a $21 \times 21$ anisotropic Gaussian blur kernel in the experimental setup. Within the SVKR, $T$ was set as 3, and $\gamma$ was set to 0.9. For monocular depth estimation, we use code from \cite{Ranftl2022, birkl2023midas}. In the context of the AMF, the hidden layer dimensions of $M_{\alpha}$ and $M_{\beta}$ were specified as 128, featuring a ReLU activation layer. AMF was incorporated into the different levels of the UNet within the Diffusion Model.

\begin{figure*}[t]
  \centering
    \includegraphics[width=\linewidth]{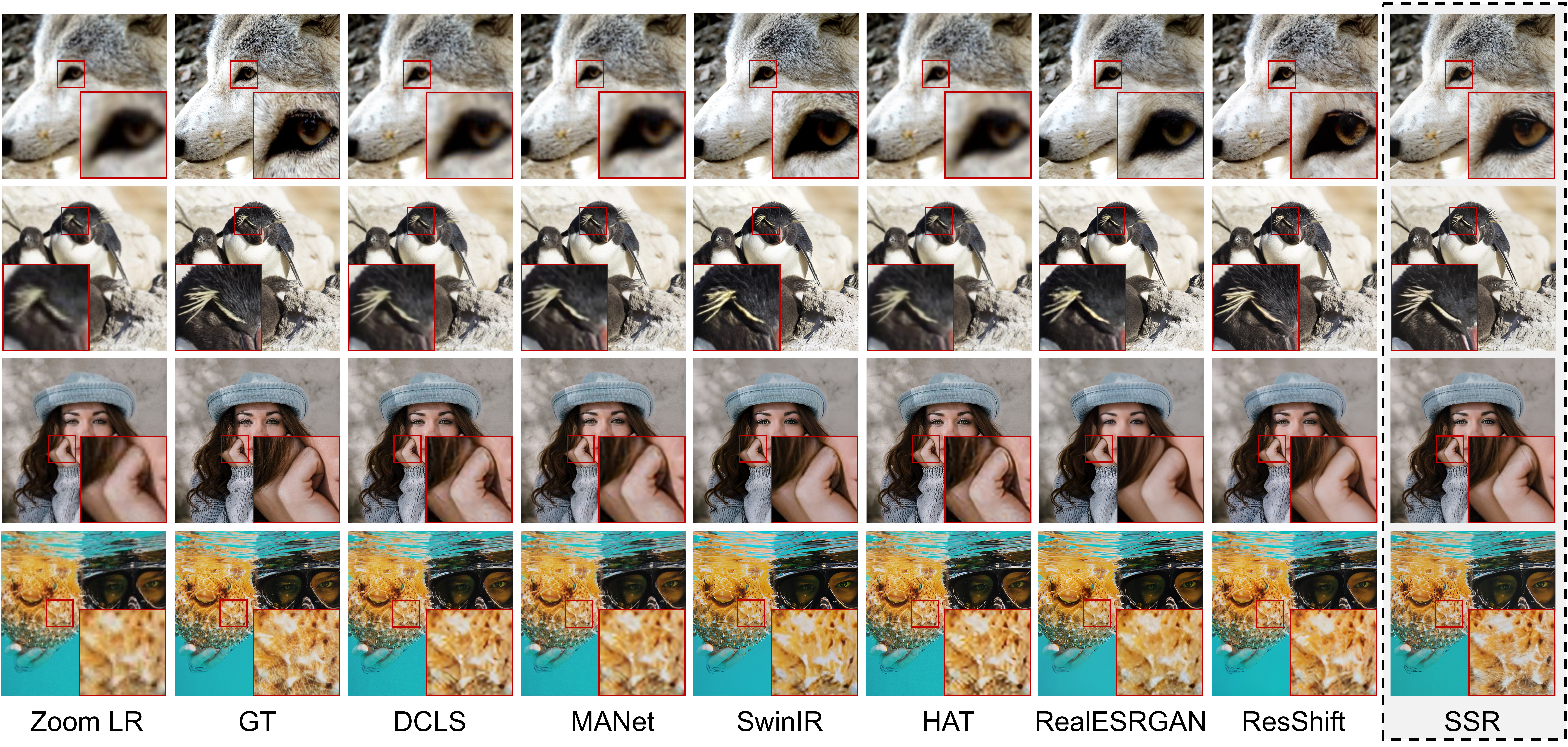}

   \caption{Visual comparisons of several representative methods on examples of the DIV2K dataset. \textbf{Zoom in for best view.}}
   \label{fig:DIV2K}
\end{figure*}
\begin{figure*}[!h]
  \centering
    \includegraphics[width=\linewidth]{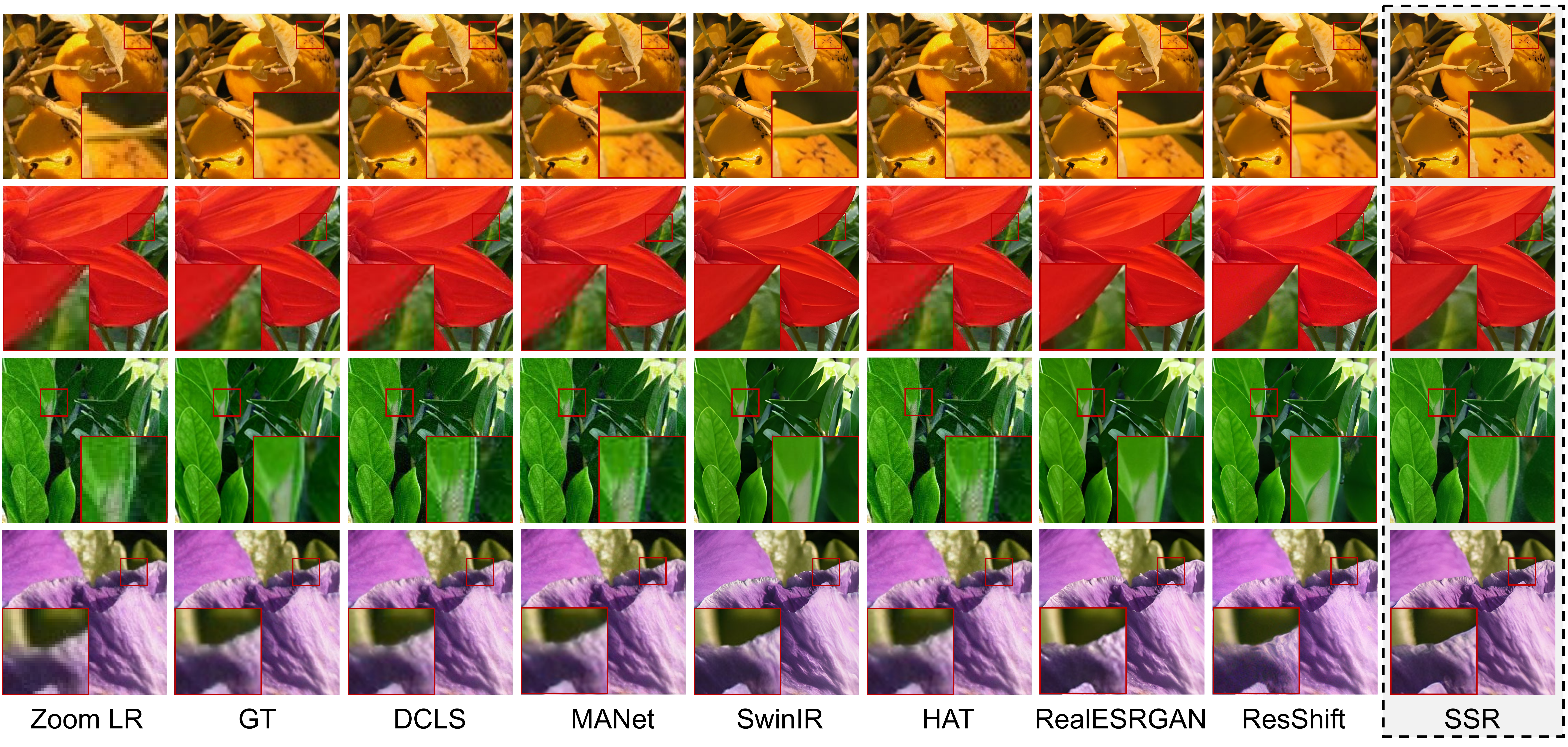}

   \caption{Visual comparisons of several representative methods on examples of the T91 dataset. \textbf{Zoom in for best view.}}
   \label{fig:T91}
\end{figure*}
\begin{figure*}[!h]
  \centering
    \includegraphics[width=\linewidth]{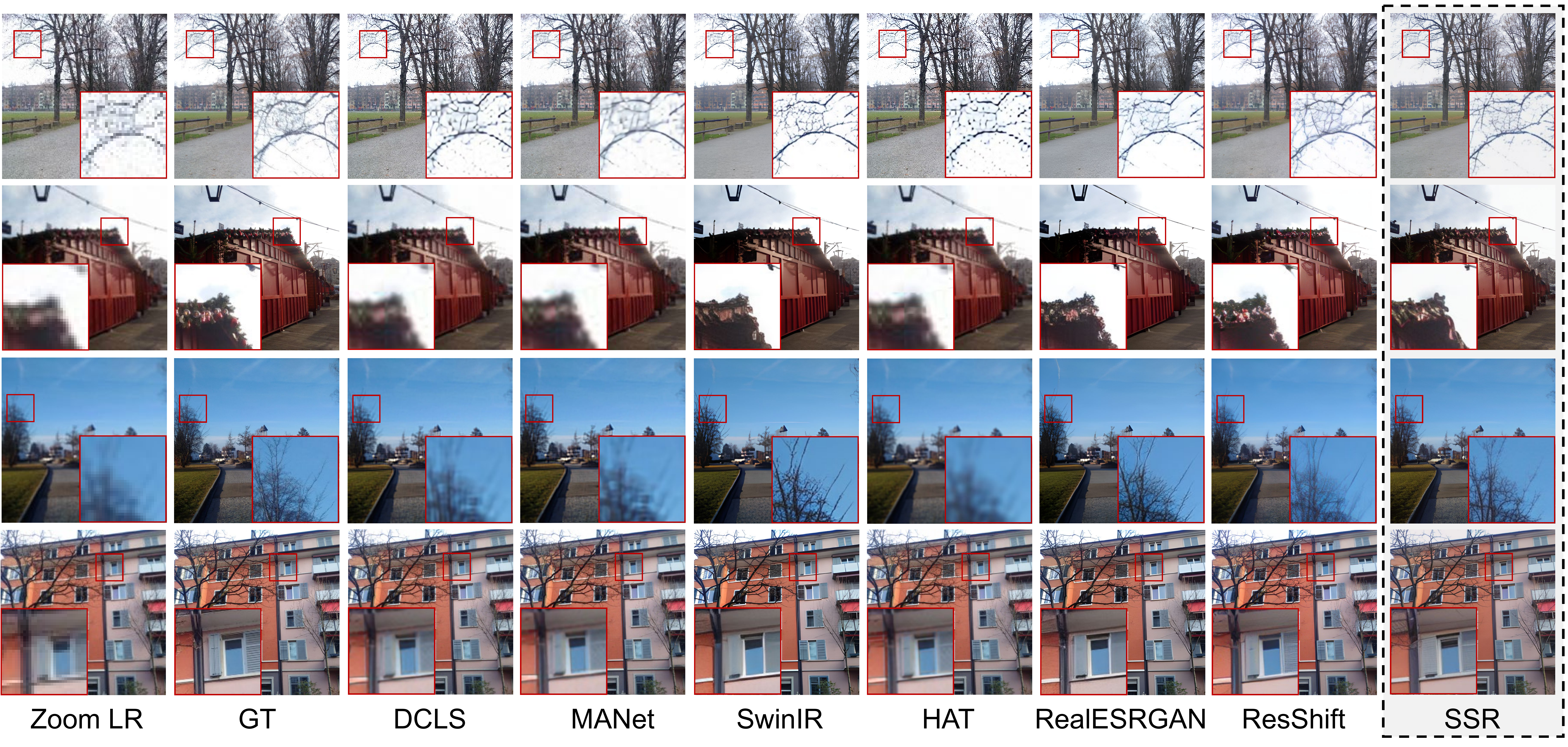}

   \caption{Visual comparisons of several representative methods on examples of the DPED-blackberry dataset. \textbf{Zoom in for best view.}}
   \label{fig:blackberry}
\end{figure*}


\noindent \textbf{Evaluation Metrics.} The PSNR and SSIM metrics are widely employed in the field of image restoration to gauge the degree of image similarity. However, it should be noted that in the realm of human perception and considering a multitude of distortions, they may fall short of faithfully representing image quality. Hence, we additionally utilize the LPIPS \cite{zhang2018perceptual}, CLIP-IQA \cite{wang2022exploring}, NIQE \cite{mittal2012making} and MUSIQ \cite{ke2021musiq} metrics to appraise the perceptual quality of the images.

\subsection{Comparison With Existing Methods}

\noindent \textbf{Quantitative Comparisons.} As Table \ref{tab: moduel} demonstrates, the performance of SSR is outstanding across multiple datasets. The experimental data reveal that SSR consistently surpasses other methods in the DIV2K, BSDS100, and Urban100 datasets. Specifically, within the DIV2K dataset, SSR achieved the highest PSNR at 22.77 dB, illustrating its exceptional capability in image quality reconstruction. Likewise, SSR secured the top spot in SSIM with a score of 0.75, signifying its precision in maintaining image structure. With an LPIPS score of 0.30, SSR significantly outperforms its competitors, with a lower score indicating superior perceived image quality. SSR also leads in the no-reference image quality assessment indices of MUSIQ, NIQE, and CLIP-IQA. Moreover, on the BSDS100 and Urban100 datasets, SSR dominates in the majority of metrics, affirming its perceptual superiority. 

\noindent \textbf{Qualitative Comparisons.} Figure \ref{fig:DIV2K}, \ref{fig:T91} and \ref{fig:blackberry} present visual comparison of the SSR methodology against established image SR techniques such as RealESRGAN, DCLS, and SwinIR. Within the experimental framework, SSR exhibits superior fidelity and an impressive capacity for detail retention, particularly in the preservation of edges and intricate textures. Previous methodologies, notably DCLS and MANet, demonstrate high sensitivity to noise within LR images, resulting in suboptimal visual outcomes—a deficiency aptly reflected in metrics like MUSIQ. SSR consistently outperforms other methods in these respects. The images generated by SSR exhibit notably higher clarity and sharpness, indicating its effectiveness in minimizing artifacts and distortions commonly introduced during the SR process. For instance, the first row of Figure \ref{fig:DIV2K}, illustrates how conventional methods grapple with blurring (DCLS, MANet), undue smoothness (Real-ESRGAN), and noise (ResShift) in the reconstruction of the eye and surrounding hair textures. SSR demonstrates a commendable balance between sharpness and naturalness, avoiding overly smoothed or exaggerated textures. 
\subsection{Ablation Study}
\noindent \textbf{Importance of the SVKR.} Our investigation initially focused on the impact of Depth-Informed Kernel (DI-Kernel), Depth Map, and SVKR on the ultimate SR outcomes. As demonstrated in Table \ref{tab: moduel}, when solely integrating either DI-Kernel or Depth Map, a significant deterioration in SSR performance was observed. However, when amalgamating information from all three modalities, the SSR's PSNR and SSIM reached 21.98 dB and 0.71, respectively. The incorporation of SVKR resulted in enhancements of 3.59\% in PSNR and 6.20\% in SSIM for the SSR. This indicates that in SSR, multimodal features synergistically supplement missing information, and SVKR effectively enhances the accuracy of information across different modalities.

\begin{table}[h]
\vspace{1mm}
\centering
\caption{Ablation studies of DI-Kernel, Depth Map and SVKR on DIV2K valid dataset. \textbf{Bold} indicates the best performance.}
\vspace{-3mm}
\scalebox{0.85}{
\begin{tabular}{ccc|ccc}
\toprule
DI-Kernel                          & \multicolumn{1}{c}{Depth Map} & \multicolumn{1}{c|}{SVKR} & PSNR $\uparrow$ & SSIM $\uparrow$ & LPIPS $\downarrow$  \\ \midrule
$\usym{1F5F8}$                     & $\usym{2613}$                 & $\usym{2613}$             & 21.26         & 0.69          & 0.32    \\
$\usym{2613}$                      & $\usym{1F5F8}$                & $\usym{2613}$             & 21.77         & 0.70          & \textbf{0.30}    \\
$\usym{1F5F8}$                     & $\usym{1F5F8}$                & $\usym{2613}$             & 21.98         & 0.71          & \textbf{0.30}    \\
\rowcolor{lightblue} $\usym{1F5F8}$ & $\usym{1F5F8}$                & $\usym{1F5F8}$            & \textbf{22.77} & \textbf{0.75} & \textbf{0.30}    \\ \bottomrule
\end{tabular}
}
\label{tab: moduel}
\vspace{1mm}
\end{table}

\begin{figure}[!h]
  \centering
    \includegraphics[width=\linewidth]{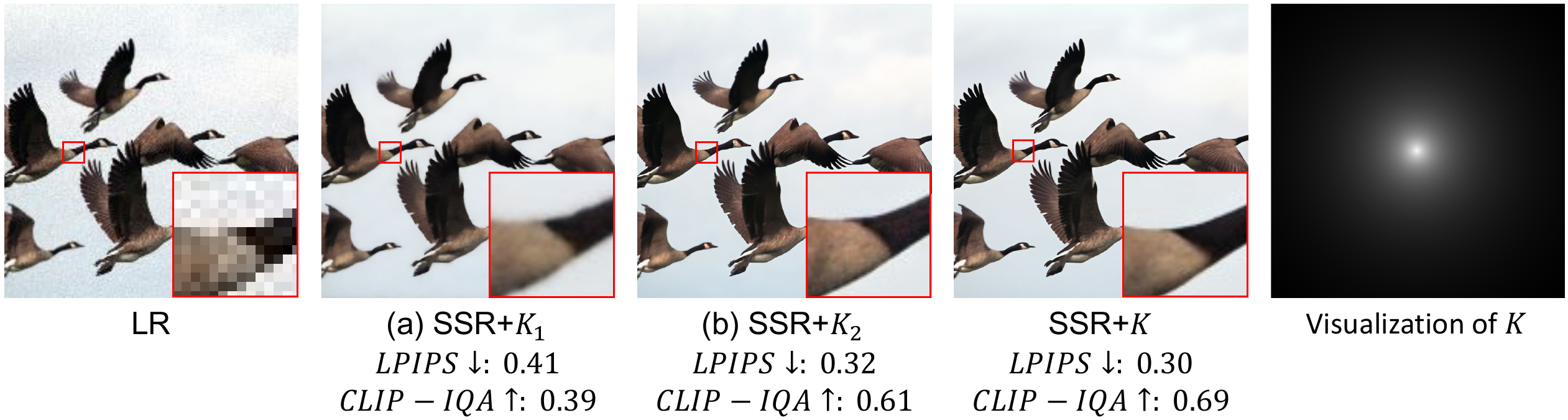}
   \caption{Ablation Comparison Between Using Accurate and Deviated Blur Kernels. (a) represents SR using $K_1$, (b) represents SR using $K_2$, and SSR denotes SR using $K$ estimated by DKENet. $K_1$ and $K_2$ respectively signify deviations from $K$ in different directions. In visualization, we calculate the variance of the blur kernel to illustrate its intensity variations and then present this data in a heatmap.}
   \label{fig:wk}
\end{figure}

\noindent \textbf{Importance of the Depth-Inforemed Kernel.} Subsequently, we examined the significance of accurately estimating the DI-Kernel. In Figure \ref{fig:wk}, SSR employs a blur kernel $K$ estimated by DKENet for image SR, where (a) and (b) represent SR using $K_1$ and $K_2$ respectively. $K_1$ and $K_2$ denote deviations from $K$ in negative and positive directions, symbolizing inaccurate blur kernels. As illustrated in Figure \ref{fig:wk}, when the blur kernel estimation is imprecise, there is a marked degradation in both visual quality and evaluation metrics of the image.

\begin{figure}[h]
  \centering
    \includegraphics[width=\linewidth]{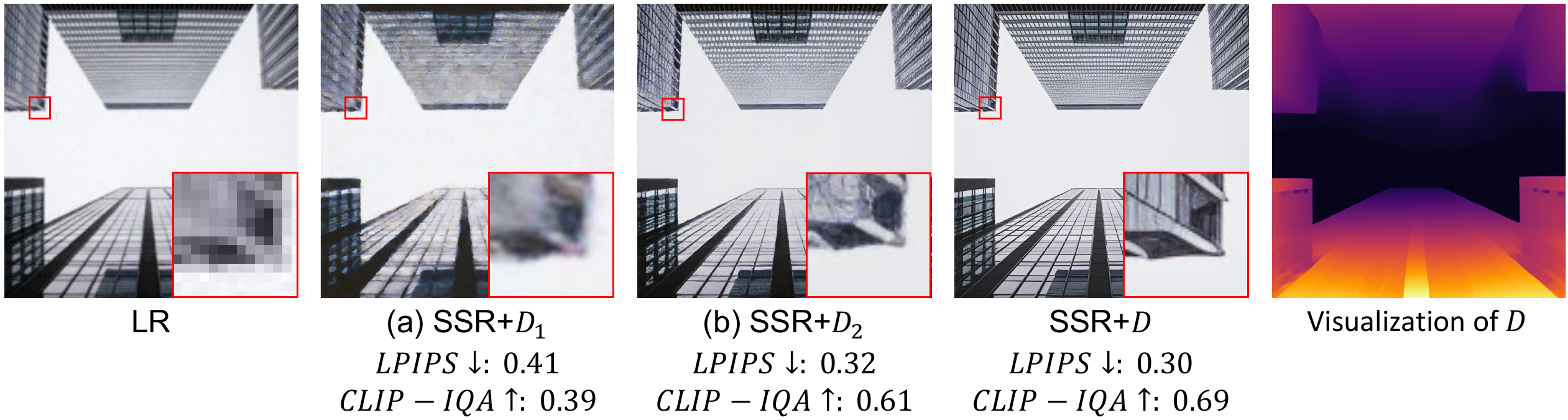}

   \caption{Ablation Comparison Between Using Accurate and Deviated Depth Map. (a) represents SR using $D_1$, (b) represents SR using $D_2$, and SSR denotes SR using $D$ estimated by Monocular depth estimate algorithm. $D_1$ and $D_2$ respectively signify deviations from $D$ in different directions.}
   \label{fig:wd}
\end{figure}

\noindent \textbf{Importance of the Depth Map.} Accurately estimating the Depth Map is also important. Figure \ref{fig:wd} shows the comparative experiments results on the Depth Map as well. Here, (a) and (b) represent deviations in the negative and positive directions, respectively, from the estimated Depth Map. The comparison in Figure \ref{fig:wd} reveals that incorrect estimations of the Depth Map can also significantly impair the performance of SSR. This emphasize the critical importance of precise estimation of Depth Map modal information.

\noindent \textbf{Performance under various degradation modes.} To validate the adaptability of SSR to image SR tasks under varying degradation modes, we conducted the experiments presented in Table \ref{tab: degradation}. These results showcase the adaptability of the image SR methodology under different degradation conditions. In scenarios devoid of noise and JPEG compression, the PSNR and SSIM metrics exhibited similar performances, demonstrating the method's consistent efficacy across diverse conditions. Overall, these outcomes affirm that the SR method can effectively adapt to a multitude of degradation modes.
\begin{table}[h]
\vspace{3mm}
\centering
\caption{Results of SSR on the DIV2K dataset under different degradation.}
\vspace{-3mm}
\scalebox{0.8}{
\begin{tabular}{ccccccc}
\toprule
Degradation  & PSNR $\uparrow$ & SSIM $\uparrow$ & LPIPS $\downarrow$ & MUSIQ $\uparrow$ & CLIP-IQA $\uparrow$ & NIQE $\downarrow$ \\ \midrule
w/o noise    & 22.17 & 0.72 & 0.42  & 56.98 & 0.39     & 6.13 \\
w/o jpeg     & 22.41 & 0.74 & 0.41  & 54.12 & 0.39     & 6.26 \\
noise + jpeg & \textbf{22.77} & \textbf{0.75} & \textbf{0.30}  & \textbf{71.44} & \textbf{0.69}     & \textbf{3.82} \\ \bottomrule
\end{tabular}
}
\label{tab: degradation}
\vspace{1mm}
\end{table}
\section{Conclusion and Discussion}
The SSR framework represents a significant advancement in the field of image SR. By integrating depth information and addressing the spatially variant in blur kernels, this framework substantially improves the authenticity of SR images. The empirical evidence from quantitative and qualitative analyses, along with ablation studies, underscores the efficacy and superiority of our approach, marking a notable contribution to the domain of image processing and SR technology. SSR also holds developmental potential in other low-level vision tasks, such as deblurring and de-jittering, which are reserved for our future work.


\clearpage
\section*{Acknowledgements}
This work was supported by National Natural Science Foundation of China (No.62072112), Scientific and Technological innovation action plan of Shanghai Science and Technology Committee (No.22511102202), National Key Research and Development Program of China (2023YFC3604802), China Postdoctoral Science Foundation under Grant (2023M730647, 2023TQ0075).

%
%
\bibliographystyle{splncs04}
\bibliography{main}
\end{document}